\documentclass[11pt,a4paper]{article}
\usepackage[hyperref]{emnlp2020}
\pdfoutput=1
\usepackage{times}
\usepackage{latexsym}
\usepackage{array,booktabs}
\usepackage{xurl}

\aclfinalcopy 

\title{Why Not Simply Translate?\\A First Swedish Evaluation Benchmark for Semantic Similarity}

\author{Tim Isbister \\
  Peltarion \\
  \texttt{tim.isbister@peltarion.com} \\\And
  Magnus Sahlgren \\
  RISE \\
  \texttt{magnus.sahlgren@ri.se} \\}

\date{}

\begin{document}
\maketitle
\begin{abstract}
	This paper presents the first Swedish evaluation benchmark for textual semantic similarity. The benchmark is compiled by simply running the English STS-B dataset through the Google machine translation API. This paper discusses potential problems with using such a simple approach to compile a Swedish evaluation benchmark, including translation errors, vocabulary variation, and productive compounding. Despite some obvious problems with the resulting dataset, we use the benchmark to compare the majority of the currently existing Swedish text representations, demonstrating that native models outperform multilingual ones, and that simple bag of words performs remarkably well.
\end{abstract}

\section{Introduction}
Semantic Textual Similarity (STS) is a
foundational concept in Natural Language Processing (NLP), with application in a wide range of tasks including Text Categorisation, Text Clustering, Text Summarisation, Recommender Systems, Information Retrieval, Question Answering, and so on. These, and other tasks, benefit from the ability to quantify the {\em semantic} similarity between two texts, $t_1$ and $t_2$. This is done by representing each text by vectors $\vec{t_1}, \vec{t_2}$ such that the similarity $sim(\vec{t_1}, \vec{t_2})$ is high if the texts are semantically similar and low if they are not. The text representations are often produced by using some type of {\em distributional} model \cite{Sahlgren2008,gastaldi}, be it word embeddings or contextualized language models.

A main challenge in research on STS is how to evaluate the quality of the text representations. The arguably most straightforward way to evaluate semantic text representations is to use manually annotated data where pairs of texts are assigned a similarity score. The objective of an STS model is then to produce similarity scores that correlate with the human judgements. This has proven to be a useful and productive approach to promote development of STS models, in particular for English, where there exist high-quality testdata. For other languages, such as Swedish, the situation is not as simple, and there is currently no publicly available evaluation data to facilitate the development of Swedish STS models. This is a major bottleneck at the moment for Swedish NLP, which needs to be resolved. This paper presents a first simple step towards Swedish STS data, pending a more measured and rigorous approach.\footnote{\url{https://www.vinnova.se/p/superlim-en-svensk-testmangd-for-sprakmodeller/}} 

\section{Data and Method}

%

\begin{table*}[t]
	\begin{center}
		\begin{tabular}{lllll}
			Language & Vocab size & Lexical richness & Avg.word length & Avg.sentence length \\
			\hline
			English  & 13~573     & 0.08             & 4.65            & 11.44               \\
			Swedish  & 19~229     & 0.12             & 5.21            & 10.73               \\
		\end{tabular}
		\caption{The English and Swedish data compared with respect to vocabulary, word length and sentence length.}
		\label{tab:dataset_1}
	\end{center}
\end{table*}

The arguably cheapest and most efficient way to produce a benchmark for semantic similarity in Swedish is to use machine translation to translate English STS data. In this paper, we use the English STS-B corpus from the GLUE benchmark,\footnote{\url{http://ixa2.si.ehu.es/stswiki}} since it is one of the standard evaluation resources for semantic similarity. 
The English STS-B data consists of sentence pairs with human similarity ratings that range from 5.00 from most similar to 0.00 for most dissimilar.
We translate the English data to Swedish using the Neural Machine Translation (NMT) model provided by Google.\footnote{\url{https://cloud.google.com/translate/docs/advanced/translating-text-v3}} 
Table \ref{tab:dataset_1} shows the vocabulary size, lexical richness (or type-token ratio), as well as average word and sentence length of the original English and translated Swedish data. Note the increase in word length in the Swedish data, which is caused by compounds. Note also the increase in vocabulary size and lexical richness, which is likely due to artefacts from the machine translation (more about this in the next Section). The Swedish dataset is publicly available and can be accessed from Github.\footnote{\url{https://github.com/timpal0l/sts-benchmark-swedish}}



\section{Error Analysis}
\label{sec:error}
There are of course a number of issues resulting from the machine translation process. One is the presence of anglicisms, where the translation is not literally incorrect, but where there exists a more conventional Swedish form. One example is the sentence ``a plane is taking off,'' which is translated to ``ett plan tar fart.'' Although it would be possible to use this construction in Swedish (the literal meaning is ``a plane takes speed''), a more conventional translation would be ``ett plan lyfter.'' The corresponding sentence pair in the STS-B data is ``a plane is taking off'' / ``an air plane is taking off,'' which in the machine translated result becomes ``ett plan tar fart'' / ``ett luftplan tar fart.'' Note the unconventional translation of ``taking off'' (``tar fart'' instead of the more conventional ``lyfter''), as well as the unconventional (but not strictly incorrect) term ``luftplan'' instead of the more conventional ``flygplan.'' Even though this sentence pair may be regarded as pragmatically incorrect from a translation perspective, it is not obvious that this sentence pair would {\em not} work as an evaluation item for semantic similarity measures; the only difference between these two sentences is the compound ``luftplan,'' which although being an unconventional (and somewhat archaic) term is not unrelated to the shorthand ``plan.'' From this perspective, a maximum similarity score (in the case of STS-B, this means a score of 5.00) seems reasonable for the Swedish translation.

This type of vocabulary discrepancy might not affect the usefulness of the data, since the vocabulary typically remains in the same domain. Another example of a translation error that does not affect the usefulness of the data is the apparent inability of the Google machine translation API to correctly translate different verb tenses. One particularly problematic case seems to be the difference between simple present tense and present progressive, as in ``peels'' versus ``is peeling,'' or ``brushes'' versus ``is brushing.'' Such tense differences are normally not preserved in the Swedish data, where only the simple present tense is retained; i.e.~both ``peels'' and ``is peeling'' are translated to ``skalar'' and not to ``håller på att skala,'' which would be the correct progressive form. This can be regarded as a translation error, but it has no effect on the result, since these sentences always have a maximum similarity score in the STS-B data. The same consideration applies to other types of translation errors, where the resulting translation is nonsensical (or at least very contrived), such as the sentence ``a person is folding a piece of paper,'' which becomes ``en person fälls ett papper'' (literally ``a person is felled a paper''), but where the incorrect translation occurs in both translated sentences. Thus, as long as the translation errors are consistent, they have a limited effect on the usefulness of the data.

The majority of the inconsistencies in the machine translated material concerns vocabulary. In order to arrive at a quantitative measure of the vocabulary issues in the translated data, we compare its vocabulary to the biggest Swedish vocabulary we could find, which is the Swedish Skipgram model trained on the Swedish CoNLL17 corpus, available at the NLPL word embedding repository.\footnote{\url{http://vectors.nlpl.eu/repository/20/69.zip}} This vocabulary contains 3~010~472 words, a substantial part of which are preprocessing errors and other noise (due to the data being collected from the Internet). 82.77\% of our test vocabulary can be found in the model. The other 17.22\% contain both nonsensical translation errors (``afaict'', ``airstrike-ärendet'', ``arrestationen'') as well as correct, but probably not very common, terms (``2006-versionen'', ``'aktiekursdetaljer'', ``antimissilförsvar''). Most of the 3~762 terms that do not occur in the NLPL vocabulary are compounds, which is perhaps not very surprising;
a well-known challenge when counting vocabulary in compounding languages is that the number of possible compounds is very large, if not infinite. This poses a significant challenge for token-based models such as word embeddings, which are dependent on a comprehensive vocabulary. Models that have the capacity to include subword units and character $n$-grams, such as FastText and models based on wordpiece/BPE encoding are much better suited to handle this challenge. We therefore hypothesise that the machine translated data will work better for comparison of subword/character-based models than for token-based ones.

\begin{table*}[!h]
	%
	%
	%
	\begin{center}
		\begin{tabular}{l|l|l|l}
																			
			\toprule
			Supervision         & Model                    & Language     & Test (sv)      \\
			\midrule
			                    & XLM-R                    & Multi        & 0.166          \\
			\cmidrule{2-4}
			                    & Word2Vec                 & sv           & 0.374          \\
			\cmidrule{2-4}
			                    & LaBSE                    & ``Agnostic'' & 0.411          \\
			\cmidrule{2-4}
			                    & KB/BERT                  & sv           & 0.419          \\

			\cmidrule{2-4}
			                    & fastText                 & sv           & 0.420          \\
			\cmidrule{2-4}
			                    & AF/BERT                  & sv           & 0.484          \\
																	
																	
			\cmidrule{2-4}
			                    & LASER                    & ``Agnostic'' & 0.704 \\
			\midrule
			NLI	(en)            & XLM-R $\leftarrow$ SBERT & Multi        & 0.697          \\									
			\midrule
			NLI (en) + STS (en) & XLM-R $\leftarrow$ SBERT & Multi        & 0.801 \\
			\midrule
			NLI (en) + STS (en) + STS (sv) & XLM-R $\leftarrow$ SBERT & Multi        & 0.808 \\
			\midrule
			STS (sv)            & TF                       & sv           & 0.406          \\
			\cmidrule{2-4}
			                    & TF-IDF                   & sv           & 0.547          \\
			%
			\cmidrule{2-4}
			                    & SVR-TF-IDF               & sv           & 0.704          \\
			\cmidrule{2-4}
			                    & AF/BERT                  & sv           & 0.714          \\
						
			\cmidrule{2-4}
			                    & FFNN-LASER               & "Agnostic"   & 0.764          \\
			\cmidrule{2-4}
			                    & KB/BERT                  & sv           & \textbf{0.825}          \\
			\bottomrule
																				
		\end{tabular}
		\caption{Results for the various representations on the datasets used in these experiments}
		\label{tab:res}
	\end{center}
\end{table*}

\section{Experiments}
Representation learning has been an enormously productive research area in recent years, with a progression from token-based embeddings to contextualized language models, which by now completely dominate representation learning for NLP. As a first application of the Swedish STS dataset, we compare a majority of the currently existing representation models for Swedish. This includes the following models:
\newline
\textbf{TF:}
The arguably simplest form of Bag-of-Words (BoW) representation based on term frequency. We collect term frequencies from the training and development data, and simply apply the frequencies to the test data.
\newline
\textbf{TF-IDF:}
BoW representation that weights term importance by the inverse document frequency. As with the TF representation, we count TF-IDF weights from the training and development data, and apply the weights to the test data. We use two versions in the supervised setting:
one where we simply apply the IDF weights to the test data using words as tokens, and another one where we feature engineer the IDF representation to contain character n-grams ranging from 1 to 5 characters. To get a fixed size vector, the element-wise difference between the n-gram vectors are used to train a supervised Support Vector Regressor.      
\newline
\textbf{Word2Vec:}
Shallow token-based language model \cite{mikolov2013efficient}. We use the Skipgram model from the NLPL repository, which we have already introduced in Section \ref{sec:error}. The vectors for sentences are obtained by averaging the embedding vector for each word.
\newline
\textbf{fastText:}
A variant of Word2Vec that considers character $n$-grams of the context words \cite{grave2018learning}. We use the CBOW model 
%
that has been trained on Common Crawl and Wikipedia.\footnote{\url{https://fasttext.cc/docs/en/crawl-vectors.html}} 
As with Word2Vec, the vectors for sentences are obtained by averaging the embedding vector for each word.
\newline
\textbf{BERT:}
Deep Transformer network trained using a masked language modeling objective (BERT stands for Bidirectional Encoding Representations from Transformers \cite{devlin-etal-2019-bert}). We include both currently existing Swedish versions of BERT; KB/BERT from the Royal Swedish Library \cite{swedish-bert}, and AF/BERT from the Swedish Public Employment Service. As sentence representation, we use the mean token representations from the last layer.\footnote{Using the CLS representation produced consistently lower results in our tests.}
\newline
\textbf{SBERT}: Uses a siamese setting where two BERT (or other types of Transformer) models are trained using Natural Language Inference (NLI) data in such a way that the training objective enforces similar representations for sentences with an entailment relation in the training data \cite{reimers-2020-multilingual-sentence-bert}. The resulting model has been demonstrated to produce useful sentence representations (hence the name Sentence-BERT, or SBERT in short) that outperform the standard BERT representations.
\newline
\textbf{XLM-R:} The RoBERTa Transformer model trained using a multilingual masked language modeling objective on massively multilingual data \cite{conneau2019unsupervised}. 
\newline
\textbf{LASER:}
Contextualized language model based on a BiLSTM encoder trained using a translation objective on parallel data (LASER stand for Language-Agnostic SEntence Representations \cite{DBLP:journals/corr/SchwenkTFD17}). The term ``agnostic'' is used because the model is claimed to handle more than 90 different languages. Since its not possible to retrain the whole architecture of the current LASER implementation, only the final representation can be used. We use this in a supervised manner by taking the element-wise difference from the embeddings and train a fully connected layer with Adam as optimizer. 
\newline
\textbf{LaBSE:} A BERT variant trained on massively multilingual data using both masked language modeling and translation language modeling objectives. The resulting model is called Language-agnostic BERT Sentence Embedding \cite{feng2020languageagnostic}.
Similarly to LASER, the term ``agnostic'' is used because the model is claimed to handle more than 100 languages.
%

For each unsupervised model, we produce a fixed-sized vector for each sentence, and compare sentence pairs using cosine similarity, and for the supervised models the regression output is used. We use Pearson correlation coefficient to compare the resulting similarity measures with the gold labels of STS-B. 

\section{Results}
%
%
%
%
Table \ref{tab:res} summarizes the results. Note that there is no consistent difference between token-based embeddings and contextualized ones, when there is no supervision for the sentence representations. In particular XLM-R underperforms in the unsupervised case, and the most recent LaBSE model does no better than fastText embeddings. 
The best model in the unsupervised setting is LASER, which seems to produce useful sentence representations for Swedish even without supervision.

Using SBERT significantly improves the performance of XLM-R, which is expected. Adding finetuning with the English STS-B data further improves the performance, and adding Swedish finetuning on top improves the result even further. This demonstrates the capacity for cross-lingual transfer using multilingual models. Adding supervision to the native Swedish models improves their performance, and our best score is reached by the KB/BERT model finetuned on the Swedish data. Note that the simple BoW model with Support Vector Regression reaches a performance of 0.704, which is remarkably competitive considering the enormous difference in computational cost between this and the other models.

\section{Conclusions}
Machine translation introduces a number of issues into the data, mostly concerning vocabulary. We argue that this is problematic for token-based models, but should be manageable for subword- and character-based models. We thus do not recommend that the machine translated STS-B data is used with standard word embeddings, but language models that rely on wordpiece/BPE tokenisation should be able to handle the vocabulary issues, and as such should be amenable to comparison using the Swedish STS-B dataset introduced in this paper.

Due to the high prevalence of translation errors, we do not recommend that the translated data is used to train or finetune models for downstream deployment. The translation errors likely have a limited effect for comparison between different models, but it is unclear what effects they might have for downstream application.

\section*{Acknowledgements}

This work was partly supported by Vinnova under grant 2019-02996. We wish to thank anonymous reviewer \#2 for valuable comments.

\bibliographystyle{acl_natbib}
\bibliography{emnlp2020,anthology}
\end{document}